\documentclass[conference]{IEEEtran}
\IEEEoverridecommandlockouts
\usepackage{cite}
\usepackage{amsmath,amssymb,amsfonts}
\usepackage{algorithmic}
\usepackage{subfigure}
\usepackage{graphicx}
\usepackage{textcomp}
\usepackage{xcolor}
\usepackage{color,soul}
\usepackage{multirow}
\def\BibTeX{{\rm B\kern-.05em{\sc i\kern-.025em b}\kern-.08em
    T\kern-.1667em\lower.7ex\hbox{E}\kern-.125emX}}
\bstctlcite{IEEEexample:BSTcontrol}

\makeatletter
\newcommand{\linebreakand}{%
  \end{@IEEEauthorhalign}
  \hfill\mbox{}\par
  \mbox{}\hfill\begin{@IEEEauthorhalign}
}
\makeatother

\begin{document}

\title{Study of Signal Temporal Logic Robustness Metrics for Robotic Tasks Optimization
}

\author{\IEEEauthorblockN{
Akshay Dhonthi\IEEEauthorrefmark{1}\IEEEauthorrefmark{2},
Philipp Schillinger\IEEEauthorrefmark{1},
Leonel Rozo\IEEEauthorrefmark{1} and
Daniele Nardi\IEEEauthorrefmark{3}}
\IEEEauthorblockA{\IEEEauthorrefmark{1}
Bosch Center for Artificial Intelligence, Renningen, Germany \\
\IEEEauthorblockA{\IEEEauthorrefmark{2},\IEEEauthorrefmark{3}
Department of Artificial Intelligence and Robotics, Sapienza University, Rome, Italy\\ }
Email: \IEEEauthorrefmark{2}dhonthirameshbabu.1887502@studenti.uniroma1.it, \IEEEauthorrefmark{1}\{philipp.schillinger, leonel.rozo\}@de.bosch.com, \\ \IEEEauthorrefmark{3}nardi@diag.uniroma1.it}}

\maketitle

\begin{abstract}
Signal Temporal Logic (STL) is an efficient technique for describing temporal constraints.
It can play a significant role in robotic manipulation, for example, to optimize the robot performance according to task-dependent metrics.
In this paper, we evaluate several STL robustness metrics of interest in robotic manipulation tasks and discuss a case study showing the advantages of using STL to define complex constraints.
Such constraints can be understood as cost functions in task optimization. 
We show how STL-based cost functions can be optimized using a variety of off-the-shelf optimizers.
We report initial results of this research direction on a simulated planar environment. 
\end{abstract}

\section{Introduction}
Many robotic manipulation tasks are sensitive to small changes in execution parameters like target positions or stiffness.
Also, applications usually benefit significantly from improved execution time or reliability.
However, it usually is a challenge to optimize such tasks due to their sequential nature.
To handle such a challenge, \emph{Signal Temporal Logic} (STL) which is a formalism to describe the temporal characteristics of trajectories is well-suited.
It provides a real-valued function, called \emph{robustness metric} generated from a logical specification which can be used to evaluate the performance of the robotic tasks.

In this work, we investigate the suitability of different robustness metrics for the purpose of optimizing robotic manipulation tasks.
There have been efforts towards applying STL in various robotic environments.
\cite{gundana2021event, lindemann2020barrier} develop control techniques for multi-agent systems using STL specification.
\cite{puranic2021learning} has used STL rewards in Learning from Demonstration (LfD) on a 2D driving scenario.
However, most works focus on directly optimizing the controller, which can be limiting for complex systems.
Instead, we propose to use classical black-box optimization to improve existing tasks.

\section{STL Robustness Metric} \label{Sec: RobMetric}
Formally, an STL specification can be understood as follows.
Consider a discrete time sequence $t := \{t_k | k \in \mathbb{Z}_{\geqslant0}\}$.
The STL formula $\varphi$ is defined using the predicate $\mu$ that is of the form $f(x(t))$, where $x(t)$ is the state of the signal at time $t$ and $f$ maps each time point to the real-value $x(t_k)\in \mathbb{R}$.
The STL syntax is defined as:
$
    \varphi \;  := \; \mu \; | \; \neg\varphi \; | \; \varphi_1\land\varphi_2 \; | \; \varphi_1\lor\varphi_2 \; | \; \mathbf{G}_{I}\varphi \; | \; \mathbf{F}_{I}\varphi \; | \; \varphi_1 \mathbf{U}_{I}\varphi_2
$. 
Where $I = [a, b]$ is the set of all $t_k \in t$ such that $a, b \in t$; $b > a \geq 0$.
The operators $\neg, \land, \lor$ refer to Boolean \emph{negation}, \emph{conjunction} and \emph{disjunction} operators respectively.
The temporal operators $\mathbf{G}, \textbf{F}, \mathbf{U}$ refer to \emph{globally}, \emph{eventually} and \emph{until} operators, respectively.
A complete definition of STL can be found in \cite{aksaray2016q},\cite{mehdipour2019arithmetic}. 

The \emph{robustness metric} denoted as $r(\varphi, x, t) \in \mathbb{R}$ is the quantitative semantics of the STL formula $\varphi$ that measures "how well" the signal $x$ is fulfilled at time $t$.
The classical way of defining this semantics is \emph{space robustness} \cite{belta2019formal}. 
This measure is positive if and only if the signal satisfies the specification (\emph{Soundness} property) and the closer the robustness is to zero, the smaller are the required changes of signal values to change the truth value.
Formally, space robustness and its corresponding operators are defined as follows. 
\begin{equation}
  \begin{aligned}
    r(\mu, x, t) =&\; f(x(t)) , \\
    r(\neg\varphi, x, t) =&\; -r(\mu, x, t) ,\\
    r(\varphi_1\land\varphi_2, x, t) =&\; \operatorname{min} (r(\varphi_1, x, t), r(\varphi_2, x, t)) ,\\
    r(\varphi_1 \mathbf{U}_{[a,b]}\varphi_2, x, t) =& \operatorname{max}_{t_{k1} \in [t+a, t+b]} (\operatorname{min} (r(\varphi_1, x, t_{k1}),\\ 
        & \operatorname{min}_{t_{k2} \in [t+a, t+t_{k1}]}r(\varphi_2, x, t_{k2}) )) .
  \end{aligned}
\label{Eqn: SpaceRobustness}
\end{equation}

The remaining operators can be derived using~\eqref{Eqn: SpaceRobustness}. 
Although this is an intuitive and practical way to determine robustness, it has limitations, particularly due to the \emph{min} and \emph{max} functions.
They are non-smooth and non-differentiable, making it more difficult for any optimizer to find a good solution.
To better address this issue, we discuss other alternatives to space robustness and some of the properties. 

\subsection{Robustness types}
To address the above issues of space robustness, several alternative definitions have been proposed in recent years.

\subsubsection{Time Robustness}
Instead of looking at how well individual signal values satisfy the specification, time robustness shifts the signal in time to quantify satisfaction \cite{donze2010robust}. 
However, time robustness is discontinuous at rising and falling edges due to the switching from positive to negative values.
This kind of robustness is useful in cases where it is important to find a signal based on how fast/slow they should satisfy the specification.

\subsubsection{LSE Robustness} \label{Sec: LSE}
To deal with the smoothness problem of space robustness, \cite{pant2017smooth} proposed a robustness formulation based on the \emph{Log-Sum-Exponential} (LSE) approximation of the \emph{max} and \emph{min} operators.
LSE corresponds to the log of a summation of exponential terms with some scalar multiplier $k \in \mathbb{R}_{\geq 0}$ as scaling factor.
This approximation is smooth due to infinitely differentiable approximation and can reduce the influence of spikes in the signal, but it comes with a cost of loosing soundness property. 

\subsubsection{AGM Robustness} \label{Sec: AGM}
The \emph{Arithmetic Geometric Mean} (AGM) is an average-based robustness that modifies all STL operators at all time instances \cite{mehdipour2019arithmetic}. 
This definition captures how fast the signal $x \in [-1,1]$ satisfies the specification by computing arithmetic and geometric means.
This normalizing approach can be useful when the signals are of different units.
However, AGM does not guarantee convergence with gradient-based optimization techniques \cite{gilpin2020smooth}.

\subsubsection{Smooth Robustness}
It can be seen as a combination of \emph{LSE} and \emph{AGM} robustness metrics \cite{gilpin2020smooth} because, unlike them, it is both sound and guarantees convergence.
The use case is similar to \emph{LSE} with additional feature of giving positive outcome only if the signal satisfies the specification.

\subsubsection{Avg Robustness}
It is a combination of \emph{space} and \emph{time} robustness.
It captures how soon or late a signal meets the specification \cite{aksaray2016q} and therefore is suitable to optimize both accuracy and speed in achieving a task.
The Avg Robustness does not support nested temporal operators, which can be a drawback. 

\subsubsection{NEW Robustness}
It is introduced recently in \cite{varnai2020robustness} which is based on a scale-invariant behaviour.
The metric is computed by taking weighted average of the effective measures
where the weight is defined such that it becomes traditional \emph{space robustness} when it approaches infinity.
The test results from the authors show better performance comparing to \emph{AGM robustness} in terms of finding feasible solutions.

\subsection{Metric Properties} \label{Sec: Properties}

\cite{varnai2020robustness} summarizes a number of properties that help to classify robustness metrics.
Since most of these properties are especially relevant for optimization, we summarize them here for all considered robustness definitions in Table ~\ref{Table: MetricComp} for the following property definitions.
A metric is \emph{sound} if $r(\varphi, x, t) \geq 0$ if and only if the signal $x$ satisfies the specification at time $t$.
A metric is \emph{weakly smooth} if it is continuous everywhere and its gradient is continuous for all points.
A metric is said to \emph{converge} if it guaranteed to reach a local maximum with gradient-based optimization.
A metric is \emph{shadow-lifting} if the metric increases when making partial progress towards the task specification.
A metric is \emph{scale-invariant} if $\operatorname{max}(\alpha a_1\land\alpha a_2) = \alpha \; \operatorname{max}(a_1\land a_2)$ for any $\alpha \in \mathbb{R}$.

\newcommand{\no}{No}
\newcommand{\yes}{Yes}

\begin{table}[t]
    \caption{Comparison of properties with Robustness Metrics}
    \begin{center}
        \begin{tabular}{|l|c|c|c|c|c|c|}
            \hline
            &\multicolumn{6}{|c|}{\textbf{Robustness Metrics}} \\
            \cline{2-7} 
            \multicolumn{1}{|c|}{\textbf{Properties}} & \textbf{\textit{Space}}& \textbf{\textit{LSE}}& \textbf{\textit{Smooth}} & \textbf{\textit{AGM}} & \textbf{\textit{Avg}} & \textbf{\textit{NEW}} \\
            \hline
            Weakly Smooth              & \no   & \yes & \yes & \yes & \no  & \yes \\
            Sound               & \yes  & \no  & \yes & \yes & \yes & \yes \\
            Converge            & \no   & \yes & \yes & \no  & \no  & \yes \\
            Shadow-lifting      & \no   & \no  & \no  & \yes & \yes & \yes \\
            Scale-invariance    & \yes  & \yes & \yes & \no  & \yes & \yes \\
            \hline
        \end{tabular}
        \label{Table: MetricComp}
    \end{center}
\end{table}

\section{Learning from STL} \label{Sec: STLSpce}
In this section, we discuss on the approach of using STL for optimizing tasks and present STL specifications to obtain a desired behavior. 
Specifically, we focus on optimizing task execution duration and final position of the robot end-effector using CMA-ES and Bayesian optimization (BO).
The optimizer gets rewards based on the STL specifications at the end of each task execution and generates different parameters for the next run to finally produce a learned trajectory that fulfills the STL constraints. 

Let us consider the following desired behavior: ``The end-effector has to \emph{eventually} visit three regions $j \in \{A, B, C\}$ within the time intervals $3$ to $4$, $8$ to $10$, and $13$ to $15$ seconds, respectively".
Equation \eqref{Eqn: STLspec} represents this as an STL specification.
\begin{equation}
\begin{aligned}
    \varphi =& \mathbf{F}_{[3,4]}\varphi_A \; \land \; \mathbf{F}_{[8,10]}\varphi_B \; \land \; \mathbf{F}_{[13,15]}\varphi_C ,\\
    \varphi_j =& x_{j,lb} < x_j < x_{j,ub} \; \land \; y_{j,lb} < y_j < y_{j,ub} .
\end{aligned}
\label{Eqn: STLspec}
\end{equation}
These specifications show the potential of STL for defining temporal constraints for task optimization.
In general, we obtain a lower reward when the behavior fails the constraints soon.
The rewards obtained using such specifications can push the optimizer to satisfy all the conditions even when those constraints are in different time instants.
This specification is simple to change to achieve different desired outcomes.

\section{Experiments and Results}
We conduct experiments on the 7-DOF Panda Robot in a simulated environment.
The number of evaluations per experiment is $60$, each one taking $25$ minutes approximately.
Time taken for reward computation and optimization is fast.
The robot motion profile has three point to point trajectories. 
The objective here is to optimize duration parameters and $(x,y)$ end-effector position parameters at each trajectory which totals to 9 parameters.
The robot resets to its initial position at the end of each evaluation.

\begin{figure}[t]
    \centering
    \includegraphics[trim={0cm 0cm 0cm 0.8cm},clip, width=.40\textwidth]{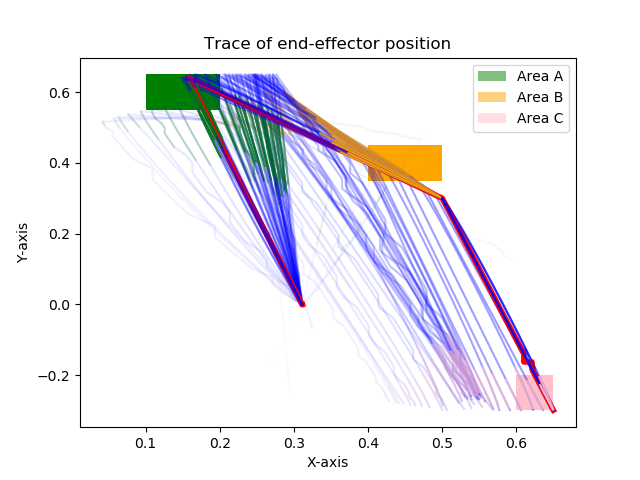}
    \caption{
    End-effector trajectories (blue solid lines) at each BO iteration using the \emph{NEW} robustness metric, where color intensity depicts the optimization evolution.
    The red line represents the highest reward execution. 
    The lines has colored patches which are the time domain when the end-effector has to reach the respective regions.
    }
    \label{fig:XYTrace}
\end{figure}

\begin{figure}[htbp]
    \centering
    \includegraphics[width=7.0cm]{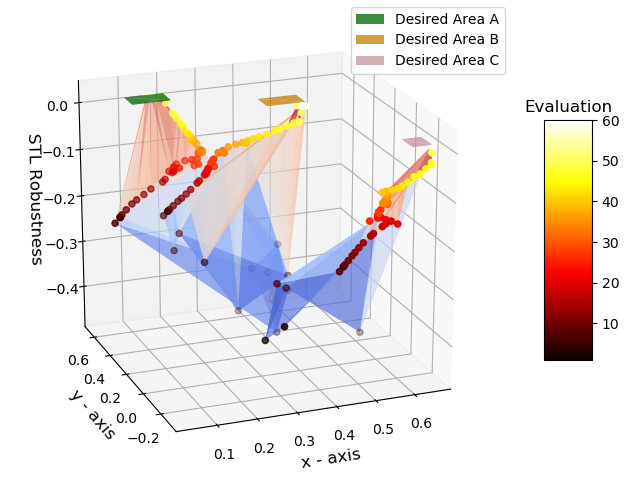}
    \caption{STL Robustness values as a function of the $(x,y)$ values provided by BO during task optimization considering the three desired regions A, B, C (depicted by colored planes). The used metric is \emph{NEW} robustness.}
    \label{fig:RewardFlow}
\end{figure}

\begin{figure}[htbp]
    \centering
    \includegraphics[width=7.0cm, trim={0 0 0 1.0cm},clip,width=.35\textwidth]{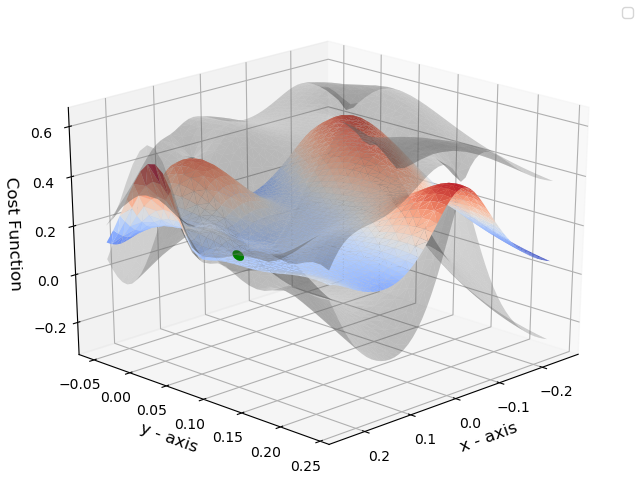}
    \caption{The cost of Gaussian process as a function of $(x, y)$ positions of the region B at the end of the experiment with \emph{Smooth} robustness. Cost here denotes negative \emph{STL Robustness}. The mean of the GP is shown as colored surface. The $\pm$ variance is given as grey surface. The green dot is the location of the minimum underlying cost.}
    \label{fig:GP}
\end{figure}

\begin{figure}[h]
    \centering
    \includegraphics[width=8.0cm, trim={0 0.4cm 0 11.6cm},clip,width=.50\textwidth]{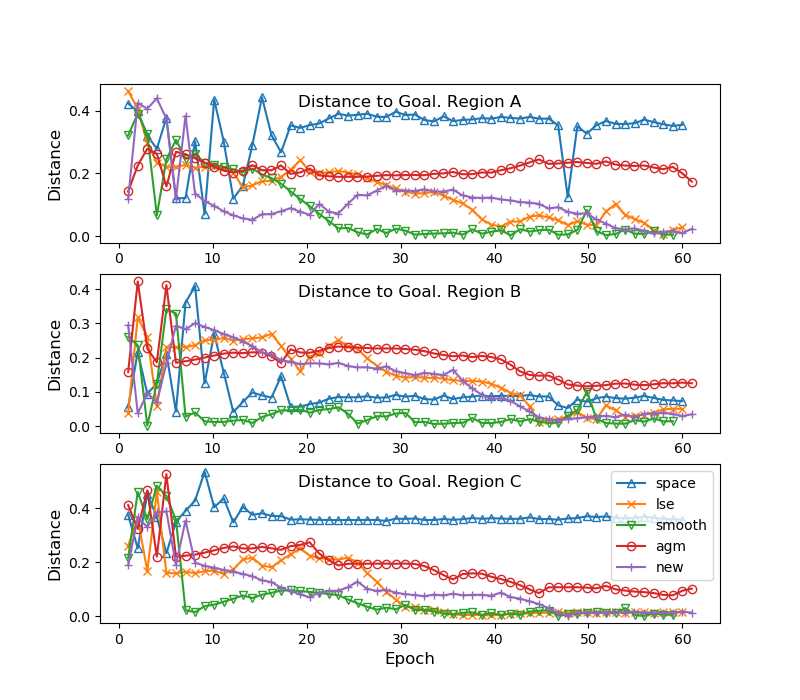}
    \caption{The figure shows at each epoch, how close the end-effector is to the goal during respective time frames for regions C. The optimizer used is BO.}
    \label{fig:DistToGoal}
\end{figure}

The trace of the end-effector in Fig.~\ref{fig:XYTrace} shows trajectory positions converging to the goal regions.
Figure~\ref{fig:RewardFlow} displays the obtained rewards, which increase when the trajectories get closer to the desired regions.
Figure~\ref{fig:GP} shows the BO surrogate model, i.e. a \emph{Gaussian Process} (GP), after evaluating $60$ iterations.
It is clear from Fig.~\ref{fig:DistToGoal} that the optimization is heavily affected by the robustness metric used.
Table~\ref{Table: ExpComp} shows the performance of all the metrics using \emph{BO} and \emph{CMA-ES} optimizers.
The \emph{Success Rate} (SR) is computed as the number of times the robot satisfies all the constraints over $60$ epochs.
The \emph{Task Satisfaction} (TS) is the first evaluation step when all the STL constraints were satisfied.
Some experiments are run until convergence.
\emph{Smooth Robustness} performs better with \emph{BO} while \emph{NEW robustness} performs better using \emph{CMA-ES}.

Based on observations from the experiments, optimization with the \emph{space robustness} metric does not always satisfy the constraints when there are more temporal operators.
\emph{LSE} and \emph{Smooth robustness} performance is influenced by their respective scaling factors and has to be tuned for different problems.
The convergence with \emph{AGM} metric is not guaranteed unless signals are normalised, but this is not straightforward in manipulators due to the complexity of obtaining workspace boundaries.
The \emph{NEW} robustness metric is suitable for defining maniputator tasks as they are robust in finding a solution all the time while \emph{Smooth robustness} can converge faster given the scaling factors are tuned. 

\newcommand{\STAB}[1]{\begin{tabular}{@{}c@{}}#1\end{tabular}}

\begin{table}[htbp]
    \caption{Overview Table for Robustness Metric properties}
    \begin{center}
        \begin{tabular}{|c|c|c|c|c|c|c|c|}
            \hline
            & &\multicolumn{6}{|c|}{\textbf{Robustness Metrics}} \\
            \cline{3-8} 
            & \multicolumn{1}{|c|}{\textbf{Type}} & \textbf{\textit{Space}}& \textbf{\textit{LSE}}& \textbf{\textit{Smooth}} & \textbf{\textit{Avg}} & \textbf{\textit{AGM}} & \textbf{\textit{NEW}} \\
            \hline
            \multirow{2}{*}{\STAB{\rotatebox[origin=c]{90}{BO}}}
            & SR  & 12.71\%  & 10.0\% & \textbf{29.9\%} & 9.92\% & 20.0\% & 27.41\%  \\
            & TS & 75 & 68  & \textbf{33} & 85 & 39 & 51 \\
            \hline
            \multirow{2}{*}{\STAB{\rotatebox[origin=c]{90}{CMA}}}
            & SR & 1.67\%  & 5.0\% & 3.51\% & 0.0\% & 3.7\% & \textbf{6.67}\%  \\
            & TS & 58  & 48  & 32  & Fail & 50   & \textbf{29} \\
            \hline
        \end{tabular}
        \label{Table: ExpComp}
    \end{center}
\end{table}

\section{Conclusion}
In this paper, we exploited STL-based constraints as cost functions to optimize simple robotic manipulation tasks.
We analyzed several STL-based cost functions and showed their influence on optimizing simple robot trajectories in multiple-target reaching tasks.
With several experiments, it is possible to see \emph{Smooth} and \emph{NEW robustness} are suitable with classical black-box optimizers such as \emph{BO} and \emph{CMA-ES}.
Further, this work can be extended by considering orientation parameters and nested temporal STL specifications on other optimizers.

\end{document}